\def\year{2019}\relax
\documentclass[letterpaper]{article} 
\usepackage{aaai19}  
\usepackage{times}  
\usepackage{helvet}  
\usepackage{courier}  
\usepackage{url}  
\usepackage{graphicx}  
\newcommand{\citet}[1]
{\citeauthor{#1}~\shortcite{#1}}
\newcommand{\citep}{\cite}

\def\tran{^\mathrm{\scriptscriptstyle T}}

\usepackage{threeparttable}
\usepackage{tabularx}
\usepackage{booktabs}
\usepackage{subfig}
\usepackage{multirow}
\usepackage{latexsym}
\usepackage{amsmath}
\usepackage{amssymb}
\usepackage{multirow}
\usepackage{pgfplots}
\usetikzlibrary{patterns}
\usepackage{algorithm}
\usepackage{algorithmic}
\usepackage{array}

\def\bc{\mathbf{c}}

\def\b1{\mathbf{1}}

\def\bh{\textbf{h}}

\def\bo{\mathbf{o}}

\def\bp{\mathbf{p}}
\def\bq{\mathbf{q}}

\def\bu{\mathbf{u}}

\allowdisplaybreaks

\title{U-Net: Machine Reading Comprehension with Unanswerable Questions}

\author{
  Fu Sun\footnotemark[2], Linyang Li\footnotemark[2], Xipeng Qiu\footnotemark[2]\thanks{\hspace{0mm} Corresponding Author.}, Yang Liu\footnotemark[3]\\
  \footnotemark[2]\hspace{0.5mm}  Shanghai Key Laboratory of Intelligent Information Processing, Fudan University\\
\footnotemark[2]\hspace{0.5mm} School of Computer Science, Fudan University\\
\footnotemark[3]\hspace{0.5mm} Liulishuo Silicon Valley AI Lab\\
  \{fsun17,lyli15,xpqiu\}@fudan.edu.cn,\hspace{1mm} yang.liu@liulishuo.com
  }
\date{}

\begin{document}
\maketitle

\begin{abstract}
Machine reading comprehension with unanswerable questions is a new challenging task for natural language processing. A key subtask is to reliably predict whether the question is unanswerable.
In this paper, we propose a unified model, called U-Net, with three important components: answer pointer, no-answer pointer, and answer verifier.
We introduce a universal node and thus process the question and its context passage as a single contiguous sequence of tokens.
The universal node encodes the fused information from both the question and passage, and plays an important role to predict whether the question is answerable and also greatly improves the conciseness of the U-Net.
Different from the state-of-art pipeline models, U-Net can be learned in an end-to-end fashion.
The experimental results on the SQuAD 2.0 dataset show that U-Net can effectively predict the unanswerability of questions and achieves an F1 score of 71.7 on SQuAD 2.0.
\end{abstract}

\section{Introduction}

Machine reading comprehension (MRC) is a challenging task in natural language processing, which requires that machine can read, understand, and answer questions about a text. Benefiting from the rapid development of deep learning techniques and large-scale benchmarks \cite{hermann2015teaching,hill2015goldilocks,rajpurkar2016squad}, the end-to-end neural methods have achieved promising results on MRC task~\cite{seo2016bidirectional,fusionnet,drqa,clark2017simple,hu2017reinforced}.
The best systems have even surpassed human performance on the Stanford Question
Answering Dataset (SQuAD) \cite{rajpurkar2016squad}, one of the most widely used MRC benchmarks.
However, one of the limitations of the SQuAD task is that each question has a correct answer in the context passage, therefore most models just need to select the most relevant text span as the answer, without necessarily checking whether it is indeed the answer to the question.

To remedy the deficiency of SQuAD, \citet{squad2.0} developed SQuAD 2.0 that combines SQuAD with new unanswerable questions.  Table \ref{tab:example} shows two examples of unanswerable questions. The new dataset requires the MRC systems to know what they don't know.

\begin{table}[tpb]
    \centering
    \begin{tabular}{l}
Article: Endangered Species Act\\ \hline
\begin{minipage}[t]{0.9\columnwidth}%
Paragraph: ``... Other legislation followed, including
the Migratory Bird Conservation Act of 1929, a \textcolor[rgb]{0.00,0.07,1.00}{1937
treaty} prohibiting the hunting of right and gray whales,
and the \textcolor[rgb]{1.00,0.00,0.00}{Bald Eagle Protection Act of 1940}. These \textcolor[rgb]{1.00,0.00,0.00}{later
laws} had a low cost to society—the species were relatively
rare—and little \textcolor[rgb]{0.00,0.07,1.00}{opposition} was raised.”%
\end{minipage}\tabularnewline\hline
Question 1: “Which laws faced significant \textcolor[rgb]{0.00,0.07,1.00}{opposition}?”\\
Plausible Answer: \textcolor[rgb]{1.00,0.00,0.00}{later laws}\\\hline
Question 2: “What was the name of the \textcolor[rgb]{0.00,0.07,1.00}{1937 treaty}?”\\
Plausible Answer: \textcolor[rgb]{1.00,0.00,0.00}{Bald Eagle Protection Act}\\\hline
    \end{tabular}
    \caption{Unanswerable Questions from SQUAD 2.0 \protect\cite{squad2.0}.}
    \label{tab:example}
\end{table}

To do well on MRC with unanswerable questions, the model needs to comprehend the question, reason among the passage, judge the unanswerability and then identify the answer span.
Since extensive work has been done on how to correctly predict the answer span when the question is answerable (e.g., SQuAD 1.1), the main challenge of this task lies in how to reliably determine whether a question is not answerable from the passage.

There are two kinds of approaches to model the answerability of a question. One approach is to directly extend previous MRC models
by introducing a no-answer score to the score vector of the answer span ~\cite{levy2017zero,clark2017simple}.
But this kind of approaches is relatively simple and cannot effectively model the answerability of a question. Another approach introduces an answer verifier to determine whether the question is unanswerable~\cite{hu2018read,tan2018know}.
However, this kind of approaches usually has a pipeline structure. The answer pointer and answer verifier have their respective models, which are trained separately.
Intuitively, it is unnecessary since the underlying comprehension and reasoning of language for these components is the same.

In this paper, we decompose the problem of MRC with unanswerable questions into three sub-tasks: answer pointer, no-answer pointer, and answer verifier. Since these three sub-tasks are highly related,
we regard the MRC with unanswerable questions as a multi-task learning problem \cite{multiDBLP:journals/ml/Caruana97} by sharing some meta-knowledge.

We propose the U-Net to
incorporate these three sub-tasks into a unified model: 1) an answer pointer to predict a candidate answer span for a question; 2) a no-answer pointer to avoid selecting any text span when a question has no answer; and 3) an answer verifier to determine the probability of the ``unanswerability'' of a question with candidate answer information.
Additionally, we also introduce a universal node and process the question and its context passage as a single contiguous sequence of tokens, which greatly improves the conciseness of U-Net.
The universal node acts on both question and passage to learn whether the question is answerable.
Different from the previous pipeline models, U-Net can be learned in an end-to-end fashion.
Our experimental results on the SQuAD 2.0 dataset show that U-Net effectively predicts the unanswerability of questions and achieves an F1 score of 72.6.

The contributions of this paper can be summarized as follows.
\begin{itemize}
  \item  We decompose the problem of MRC with unanswerable questions into three sub-tasks and combine them into a unified model, which uses the shared encoding and interaction layers. Thus, the three-tasks can be trained simultaneously in an end-to-end fashion.
  \item We introduce a universal node to encode the common information of the question and passage. Thus, we can use a unified representation to model the question and passage, which makes our model more condensed.
  \item U-Net is very easy to implement yet effective.
\end{itemize}

\section{Proposed Model}

\begin{figure}[t]    \centering
    \includegraphics[width=0.9\linewidth]{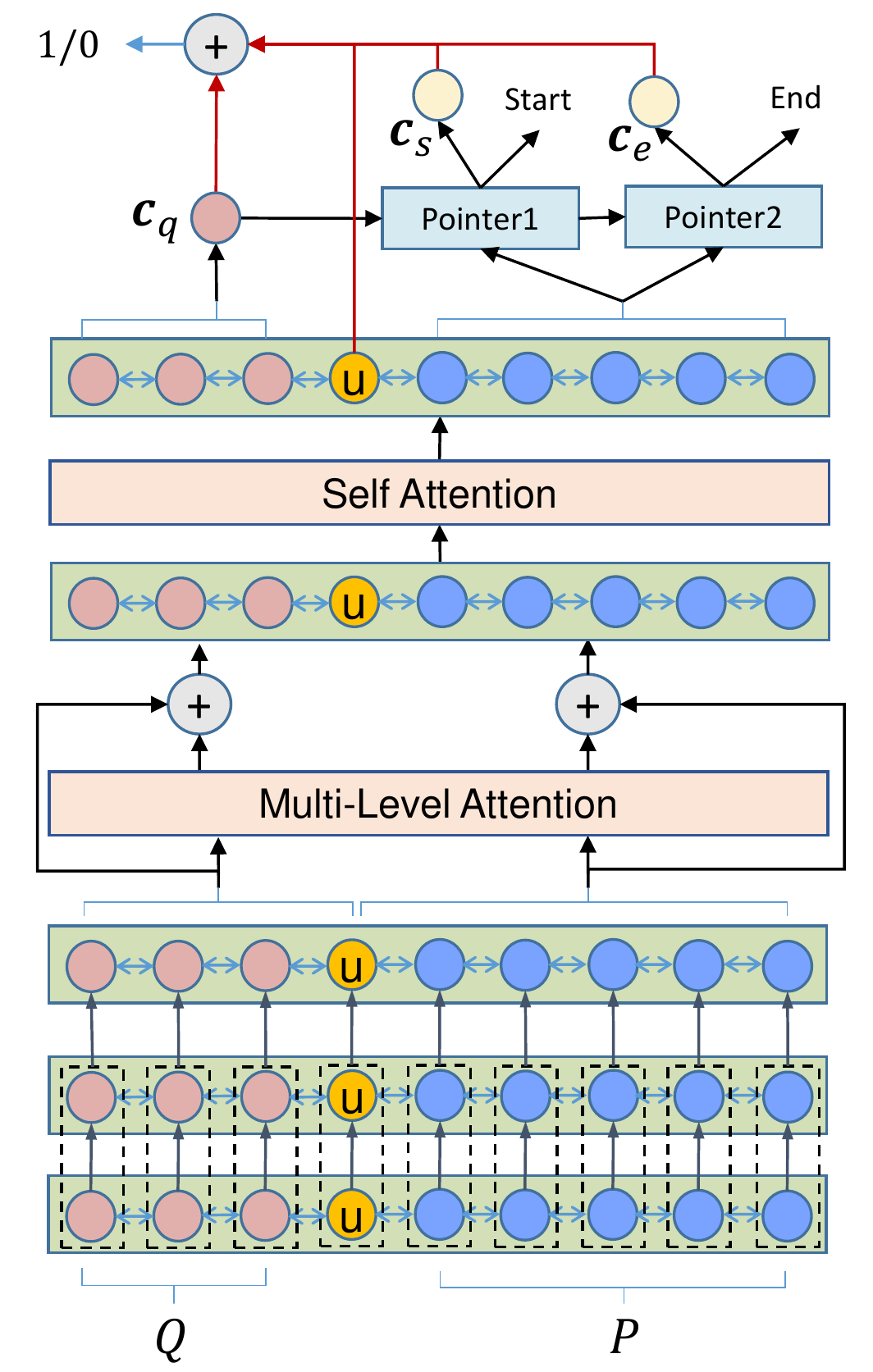}
    \caption{Architecture of the U-Net.}
    \label{fig:overview}
\end{figure}

Formally, we can represent the MRC problem as: given a set of tuples $(Q,P,A)$, where $Q = (q_1,q_2,\cdots,q_m)$ is the question with $m$ words, $P=(p_1,p_2,\cdots,p_n)$ is the context passage with $n$ words, and $A = p_{r_s:r_e}$ is the answer with $r_s$ and $r_e$ indicating the start and end points, the task is to estimate the conditional probability $P(A|Q,P)$.

The architecture of our proposed U-Net is illustrated in Figure \ref{fig:overview}.


U-Net consists of four major blocks: Unified Encoding, Multi-Level Attention, Final Fusion, and Prediction. As shown in Figure \ref{fig:overview}, we first combine the embedded representation of the question and passage with a universal node $u$ and pass them through a BiLSTM to encode the whole text. We then use the encoded representation to do the information interaction. Then we use the encoded and interacted representation to fuse the full representation and feed them into the final prediction layers to do the multi-task training.

We will describe our model in details in the following.

\subsection{(A) Unified Encoding}
\label{sub:Encoder layer}

\paragraph{Embedding} Following the successful models on SQuAD 1.1, we first embed both the question and the passage with the following features. Glove embedding  \cite{pennington2014glove} and Elmo embedding \cite{elmo_} are used as basic embeddings. Besides, we use POS embedding,  NER embedding, and a feature embedding that includes the exact match, lower-case match, lemma match, and a TF-IDF feature~\cite{drqa}. 

Now we get the question representation $Q = \bq_{i=1}^m$ and the passage representation $P = \bp_{i=1}^n$, where each word is represented as a $d$-dim embedding by combining the features/embedding described above.

\paragraph{Universal Node}
We create a universal node $u$, which is a key factor in our model and has several roles in predicting the unanswerability of question $Q$.

We expect this node to learn universal information from both passage and question.
This universal node is added and connects the passage and question at the phase of embedding, and then goes along with the whole representation, so it is a key factor in information representation.
Since the universal node is in between and later shared between passage and question,  it has an abstract semantic meaning rather than just a word embedding.

Also, the universal node is later shared in the attention interaction mechanism and used in both the answer boundary detection and classification tasks, so this node carries massive information and has several important roles in our whole model construction.

The universal node $u$ is first represented by a $d$-dim randomly-initialized vector.
We concatenated question representation, universal node representation, passage representation together as:
\begin{align}
    V=[Q,\bu, P] &= [\bq_1, \bq_2 \dots \bq_m, \bu, \bp_1, \bp_2, \cdots, \bp_n],
\end{align}
$V\in \mathbb{R}^{d \times(m+n+1)}$ is a joint representation of question and passage.

\paragraph{Word-level Fusion}
Then we first use two-layer bidirectional LSTM (BiLSTM) \cite{hochreiter1997long} to fuse the joint representation of question, universal node, and passage.
\begin{align}
    H^l &= \mathrm{BiLSTM}(V),\\
    H^h &= \mathrm{BiLSTM}(H^l),
\end{align}
where $H^l$ is the hidden states of the first BiLSTM, representing the low-level semantic information, and $H^h$ is the hidden states of the second BiLSTM, representing the high-level semantic information.

Finally, we concatenate $H^l$ and $H^h$  together and pass them through the third BiLSTM and obtain a full representation $H^f$.
\begin{align}
    H^f &= \mathrm{BiLSTM}([H^l;H^h]).
\end{align}

Thus, $H=[H^l;H^h;H^f]$ represents the deep fusion information of the question and passage on word-level.
When a BiLSTM is applied to encode representations, it learns the semantic information bi-directionally.
Since the universal node $u$ is between the question and passage, its hidden states $\bh_{m+1}$ can learn both question and passage information.
When the passage-question pair was encoded as a unified representation and information flows via the BiLSTM, the universal node has an important role in information representation.

\subsection{(B) Multi-Level Attention}
\label{sub:attention_layer}

To fully fuse the semantic representation of the question and passage, we use the attention mechanism \cite{bahdanau2014neural} to capture their interactions on different levels.

We expected that we could simply use self-attention on the encoded representation $H$ for interaction between question and passage, which contains both bi-attention \cite{seo2016bidirectional}  and self-attention \cite{selfmatchDBLP:conf/acl/WangYWCZ17} of the question and passage. But we found that it performed slightly worse than the traditional bi-directional attention with the universal node included. Therefore, we use a bi-directional attention between the question and passage.

We first divide $H$ into two representations: attached passage $H_q$ and attached question $H_p$,
and let the universal node representation $\bh_{m+1}$ attached to both the passage and question,
i.e.,
\begin{align}
    H_q &= [\bh_1, \bh_2, \cdots, \bh_{m+1}],\\
    H_p &= [\bh_{m+1}, \bh_{m+2}, \cdots, \bh_{m+n+1}],
\end{align}
Note $\bh_{m+1}$ is shared by $H_q$ and $H_p$.
Here the universal node works as a special information carrier, and both passage and question can focus  attention information on this node so that the connection between them is closer than a traditional bi-attention interaction.

Since both $H_q=[H^l_q;H^h_q;H^f_q]$ and $H_p=[H^l_p;H^h_p;H^f_p]$ are concatenated by three-level representations, we followed previous work FusionNet \cite{fusionnet} to construct their iterations on three levels.

Take the first level as an example. We first compute the affine matrix of $H^l_q$  and $H_p^l$ by
\begin{align}
    S &= \Big(\mathrm{ReLU}(W_1 H_q^l)\Big)\tran \mathrm{ReLU}(W_2 H^l_p),
\end{align}
where $S\in \mathbb{R}^{(m+1) \times(n+1)}$; $W_1$ and $W_2$ are learnable parameters.
Next, a bi-directional attention is used to compute the interacted representation $\widehat{H_q^l}$  and $\widehat{H_p^l}$.
\begin{align}
    \widehat{H_q^l} &=  H^l_p \times \mathrm{softmax}(S\tran),\\
    \widehat{H_p^l} &=  H^l_q \times \mathrm{softmax}(S),
\end{align}
where $\mathrm{softmax}(\cdot)$ is column-wise normalized function.

We use the same attention layer to model the interactions for all the three levels, and
get the final fused representation  $\widehat{H^l}, \widehat{H^h},  \widehat{H^f}$ for the question and passage respectively.

Note that while dealing with the attention output of the universal node, we added two outputs from passage-to-question attention and question-to-passage attention. So after the interaction, the fused representation  $\widehat{H^l}, \widehat{H^h},  \widehat{H^f}$ still have the same length as the encoded representation $H^l$, $H^h$ and $H^f$.

\subsection{(C) Final Fusion}

After the three-level attentive interaction, we generate the final fused information for the question and passage. We concatenate all the history information: we first concatenate the encoded representation $H$ and the representation after attention $\widehat{H}$ (again, we use $H^l,  H^h,  H^f$, and $\widehat{H^l},  \widehat{H^h},  \widehat{H^f}$ to represent 3 different levels of representation for the two previous steps respectively).

Following the success of DenseNet~\cite{densenetDBLP:journals/corr/HuangLW16a}, we concatenate the input and output of each layer as the input of the next layer.

First, we pass the concatenated representation $H$ through a BiLSTM to get $H^A$.
\begin{align}
    H^A &= \mathrm{BiLSTM}\Big([H^l; H^h; H^f; \widehat{H^l}; \widehat{H^h}; \widehat{H^f}]\Big),
    \end{align}
where the representation $H^A$ is a fusion of information from different levels.

Then we concatenate the original embedded representation $V$ and $H^A$ for better representation of the fused information of passage,  universal node, and question.
\begin{align}
    A &= [V; H^A].
\end{align}

Finally, we use a self-attention layer to get the attention information within the fused information.
The self-attention layer is constructed the same way as \cite{attentionisallyouneed}:
\begin{align}
    \widehat{A} = A \times \mathrm{softmax}(A\tran A),
\end{align}
where $\widehat{A}$ is the representation after self-attention of the fused information $A$.
Next we concatenated representation $H^A$ and $\widehat{A}$ and pass them through another BiLSTM layer:
\begin{align}
O=\mathrm{BiLSTM}[H^A; \widehat{A}].
\end{align}

Now $O$ is the final fused representation of all the information.
At this point, we divide $O$ into two parts: $O^P$, $O^Q$, representing the fused information of the question and passage respectively.
\begin{align}
    O^P &= [\bo_1, \bo_2, \cdots, \bo_{m}],\\
    O^Q &= [\bo_{m+1}, \bo_{m+2}, \cdots, \bo_{m+n+1}],
\end{align}
Note for the final representation, we attach the universal node only in the passage representation $O^P$. This is because we need the universal node as a focus for the pointer when the question is unanswerable.
These will be fed into the next decoder prediction layer.

\subsection{(D) Prediction}
\label{sub:prediction}

The prediction layer receives fused information of passage $O^P$ and question $O^Q$, and tackles three prediction tasks: (1) answer pointer, (2) no-answer pointer and (3) answer verifier.

First, we use a function shown below to summarize the question information $O^Q$ into a fixed-dim representation $\bc_q$.

\begin{align}
    \bc_q &= \sum_i \frac{\exp(W_q^\top o^Q_i)}{\sum_j \exp(W^\top o^Q_j)} o^Q_i,\label{eq:c_q}
\end{align}
where $W_q$ is a learnable weight matrix and $o_i^Q$ represents the $i_{th}$ word in the question representation. Then we feed $\bc_q$ into the answer pointer to find boundaries of answers \cite{matchDBLP:journals/corr/WangJ16a}, and the classification layer to distinguish whether the question is answerable.

\paragraph{(i) Answer Pointer}
We use this answer pointer to detect the answer boundaries from the passage when the question is answerable (i.e., the answer is a span in the passage). This layer is a classic pointer net structure \cite{pointernet}.  We use two trainable matrices $W_s$  and $W_e$ to estimate the probability of the answer start and end boundaries of the $i_{th}$ word in the passage, $\alpha_i$  and $ \beta_i$.
\begin{align}
    \alpha_i &\propto \exp(\bc_q W_s o^P_i), \\
    \beta_i &\propto \exp(\bc_q W_eo^P_i),
\end{align}

Note here when the question is answerable, we do not consider the universal node in answer boundary detection, so we have $i >0$ ($i=0$ is the universal node in the passage representation).
The loss function for the answerable question pairs is:
\begin{align}
    \mathcal{L}_{A} = -\big(\log \alpha_a +\log \beta_b \big),
\end{align}
where $a$ and $b$ are the ground-truth of the start and end boundary of the answer.

\paragraph{(ii) No-Answer Pointer}
Then we use the same pointer for questions that are not answerable.
Here the loss $\mathcal{L}_{NA}$ is:

\begin{align}
    \mathcal{L}_{NA} = -\big(\log \alpha_0 +\log \beta_0 \big),
\end{align}
$\alpha_0$ and $\beta_0$ correspond to the position of the universal node, which is at the front of the passage representation $O_p$.
For this scenario, the loss is calculated for the universal node.

Additionally, since there exits a plausible answer for each unanswerable question in SQuAD 2.0, we introduce an auxiliary \textit{plausible answer pointer} to predict the boundaries of the plausible answers. The plausible answer pointer has the same structure as the answer pointer, but with different parameters.
Thus, the total loss function is:
\begin{align}
    \mathcal{L}_{NA} = -\big(\log \alpha_0 +\log \beta_0 \big) -\big(\log \alpha'_{a^*} +\log \beta'_{b^*} \big),
\end{align}
where $\alpha'$ and $\beta'$ are the output of the plausible answer pointer; $a^*$ and $b^*$ are the start and end boundary of the unanswerable answer.

The no-answer pointer and plausible answer pointer are removed at test phase.

\paragraph{(iii) Answer Verifier}
\label{sub:Classification}
We use the answer verifier to distinguish whether the question is answerable.

Answer verifier applies a weighted summary layer to summarize the passage information into a fixed-dim representation $\bc_q$ (as shown in Eq.\eqref{eq:c_q}).

And we use the weight matrix obtained from the answer pointer to get two representations of the passage.
\begin{align}
    \bc_s &= \sum_i \alpha_i \cdot o^P_i\\
    \bc_e &= \sum_i \beta_i \cdot o^P_i
\end{align}

Then we use the universal node $\bo_{m+1}$ and concatenate it with the summary of question and passage to make a fixed vector
\begin{align}
    F= [\bc_q ;  \bo_{m+1} ;  \bc_s ;  \bc_e].
\end{align}
This fixed $F$ includes the representation $\bc_q$ representing the question information,
and $\bc_s$ and $\bc_e$ representing the passage information.
Since these representations are highly summarized specially for classification, we believe that this passage-question pair contains information to distinguish whether this question is answerable.
In addition, we include the universal node as a supplement. Since the universal node is pointed at when the question is unanswerable and this node itself already contains information collected from both the passage and question during encoding and information interaction, we believe that this node is important in distinguishing whether the question is answerable.

Finally, we pass this fixed vector $F$ through a linear layer to obtain the prediction whether the question is answerable.
\begin{align}
    p^{c} = \sigma(W_f\tran F)
\end{align}
where $\sigma$ is a sigmoid function, $W_f$ is a learnable weight matrix.


Here we use the cross-entropy loss in training.
\begin{align}
    \mathcal{L}_{AV} = -\Big(\delta \cdot\log {p^c} + (1-\delta)\cdot(\log{(1-p^c)})\Big),
\end{align}
where  $\delta\in \{0,1\}$ indicates whether the question has an answer in the passage.

Compared with other relatively complex structures developped for this MRC task, our U-Net model passes the original question and passage pair through embedding and encoding, which then interacts with each other, yielding fused information merged from all the levels.
The entire architecture is very easy to construct. After we have the fused representation of the question and passage, we pass them through the pointer layer and a fused information classification layer in a multi-task setup.

\section{Training}

We jointly train the three tasks by combining the three loss functions.
The final loss function is:
\begin{align}
    \mathcal{L} = \delta\mathcal{L}_{A}+ (1-\delta) \mathcal{L}_{NA} + \mathcal{L}_{AV},
\end{align}
where $\delta\in \{0,1\}$ indicates whether the question has an answer in the passage, $\mathcal{L}_{A}$, $\mathcal{L}_{NA}$  and $\mathcal{L}_{AV}$    are the three loss functions of the answer pointer, no-answer pointer, and answer verifier.

Although the three tasks could have different weights in the final loss function and be further fine-tuned after joint training, here we just consider them in the same weight and do not fine-tune them individually.

At the test phase, we first use the answer pointer to find a potential answer to the question, while the verifier layer judges whether the question is answerable. If the classifier predicts the question is unanswerable, we consider the answer extracted by the answer pointer as plausible.
In this way, we get the system result.

\begin{table*}[htpb]\setlength{\tabcolsep}{8pt}
    \centering
    \begin{tabular}{lcccc}
        \toprule
        \multirow{2}*{\bfseries Model} &
        \multicolumn{2}{c}{\bfseries Dev} & \multicolumn{2}{c}{\bfseries Test}\\ \cline{2-5} &EM&F1&EM&F1\\
        \midrule
        \textbf{End-to-end Model}\\
        BNA$^\star$ \cite{squad2.0}   & 59.8 & 62.6&59.2&62.1\\
        DocQA \cite{squad2.0} &65.1&67.6&63.4&66.3\\
        FusionNet++ \cite{fusionnet} &-&-&66.6&69.6\\
        SAN \cite{sanDBLP:journals/corr/abs-1712-03556}&-&-&68.6&71.4\\
        VS$^3$-Net&-&-&68.4&71.3\\
        U-Net&70.3&74.0&\textbf{69.2}&\textbf{72.6}\\
        \midrule
        \textbf{Ensemble Model}\\
        FusionNet++ (ensemble) &-&-&70.3&	72.6\\
        SAN (ensemble) &-&-&71.3	& 73.7\\
        U-Net (ensemble) &-&-&\textbf{71.5}&\textbf{75.0}\\
        \midrule
        \textbf{Pipeline Model}\\
        RMR+ELMo+Verifier \cite{hu2018read} &72.3&74.8&71.7&74.2\\
        \midrule
        Human & 86.3&89.0&86.9&89.5\\
        \bottomrule

    \end{tabular}
    \caption{Evaluation results on the SQuAD 2.0 (extracted on Sep 9, 2018). $^\star$ means the BiDAF \protect\cite{seo2016bidirectional} with No Answer.
}
    \label{tab:main}
\end{table*}

\section{Experiment}
\subsection{Datasets}
\label{sub:Datasets}
 Recently, machine reading comprehension and question answering have progressed rapidly, owing to the computation ability and publicly available high-quality datasets such as SQuAD. Now new research efforts have been devoted to the newly released answer extraction test with unanswerable questions, SQuAD 2.0 \cite{squad2.0}.
It is constructed by combining question-answer pairs selected from SQuAD 1.0 and newly crafted unanswerable questions. These unanswerable questions are created by workers that were asked to pose questions that cannot be answered based on the paragraph alone but are similar to the answerable questions. It is very difficult to distinguish these questions from the answerable ones.
We evaluate our model using this data set.
It contains over 100,000+ questions on 500+ wikipedia articles.

 \subsection{Implementation Details}
\label{sub:Implementation Details}

We use Spacy to process each question and passage to obtain tokens, POS tags, NER tags and lemmas tags of each text.  We use 12 dimensions to embed POS tags, 8 for NER tags \cite{drqa}. We use 3 binary features: exact match, lower-case match and lemma match between the question and passage  \cite{featDBLP:journals/corr/LeeKP016}.
We use 100-dim Glove pretrained word embeddings and 1024-dim Elmo embeddings.  All the LSTM blocks are bi-directional with one single layer. We set the hidden layer dimension as 125, attention layer dimension as 250.
We added a dropout layer over all the modeling layers, including the embedding layer, at a dropout rate of 0.3 \cite{dropoutDBLP:journals/jmlr/SrivastavaHKSS14}. We use Adam optimizer with a learning rate of  0.002 \cite{admaDBLP:journals/corr/KingmaB14}.

During training, we omit passage with over 400 words and question with more than 50 words.
For testing, when the passage has over 600 words and the question is over 100 words,
we simply label these questions as unanswerable.

\subsection{Main Results}
\label{sub:Main Results}

Our model achieves an F1 score of 74.0 and an EM score of 70.3 on the development set, and an F1 score of 72.6 and an EM score of 69.2 on Test set\footnote{\url{https://rajpurkar.github.io/SQuAD-explorer/}}, as shown in Table \ref{tab:main}.
Our model outperforms most of the previous approaches.
Comparing to the best-performing systems, our model has a simple architecture and is an end-to-end model.
In fact, among all the end-to-end models, we achieve the best F1 scores.
We believe that the performance of the U-Net can be boosted with an additional post-processing step to verify answers using approaches such as \cite{hu2018read}.

\subsection{Ablation Study}
\label{sub:Ablation Study}

We also do an ablation study on the SQuAD 2.0 development set to further test the effectiveness of different components in our model.
In Table \ref{tab:configuration}, we show four different configurations.

\begin{table}[htpb]
    \centering
    \begin{tabular}{lcccc}
        \toprule
        {\bfseries Configuration} & {\bfseries EM} & {\bfseries F1}& {\bfseries $\Delta$EM} & {\bfseries $\Delta$ F1} \\
        \midrule
        U-Net & 70.3& 74.0&-&-\\
        \midrule
        no node U & 67.9&71.4&-2.4&-2.6\\
        no share U & 69.7 & 73.5&-0.6&-0.5\\
        no concatenate P \& Q & 69.0 & 72.8&-1.3&-1.2\\             
        no plausible answer pointer & 69.6 & 72.9 &-0.7&-1.1\\
        no classification & 63.5 & 68.5&-6.8&-5.5\\   
        \midrule
        Self-Attn Only &69.7&73.5&-0.5&-0.5\\

        \bottomrule

    \end{tabular}
    \caption{Comparison of different configurations for our U-Net model. }
    \label{tab:configuration}
\end{table}

First, we remove the universal node $U$.  We let the negative examples focus on the plausible answer spans instead of focusing on the universal node $U$.
This results in a loss of 2.6\% F1 score on the development set, showing that the universal node $U$ indeed learns information about whether the question is answerable.

We also tried to make the universal node $U$ only attached to the passage representation when passing the attention layer.  Our results showed that when node $U$ is shared, as it is called  `universal', it learns information interaction between the question and passage, and when it is not shared, the performance slightly degraded.

As for the approaches to encode the representations, we pass both the question and passage through a shared BiLSTM. To test the effectiveness of this, we ran the experiment using separate BiLSTMs on embedded question and passage representations. Results show that the performance dropped slightly, suggesting sharing BiLSTM is an effective method to improve the quality of the encoder.

After removing the plausible answer pointer, the performance also dropped, indicating the plausible answers are useful to improve the model even though they are incorrect.

After removing the answer verifier, the performance dropped greatly, indicating it is vital for our model.

Lastly, we run a test using a more concise configuration. In the second block (multi-level attention) of the U-Net, we do not split the output of the encoded presentation and let it pass through a self-attention layer. The bidirectional attention is removed. In this way, our model uses only one unified representation of the question and passage at all time.  We simply pass this representation layer by layer to get the final result. Compared to the bi-attention model, the F1-score decreases 0.5\%.

\subsection{Multi-task Study}
\label{sub:multi-task Study}

We also run an experiment to test the performance of our multi-task model. We select different losses that participate in the training procedure to observe the performance affected by answer boundary detect or classification.

Table~\ref{tab:multi-task} shows the performance.
Here we use $EM^*$ and $F1^*$ to represent the EM and F1 score when the classification is not part of the task, which makes it very much like the task in SQuAD 1.1.

\begin{table}[htpb]
    \centering
    \begin{tabular}{lccc}
        \toprule
        {\bfseries Loss} & {\bfseries EM$^*$} & {\bfseries F1$^*$}& {\bfseries Classification Acc.}\\
        \midrule
        $\mathcal{L}$ & 75.3 & 84.8&80.2\\
        $\mathcal{L}_{AV}$&-& -&67.1\\
        $\mathcal{L}_A$ &77.2&85.1&-\\
        \bottomrule
    \end{tabular}

    \caption{Multi-task performance on the development set. }
    \label{tab:multi-task}
\end{table}

To test our classifier performance, we do not use backward propagation over the loss of answer boundary detection and simply run a classification task.
Results (the first two rows in Table~\ref{tab:multi-task}) show that there is a large gain when using the multi-task model.
The answer boundary detection task helps the encoder learn information between the passage and question and also feed information into the universal node, therefore we can use a summarized representation of the passage and question as well as the universal node to distinguish whether the question is answerable, i.e., help improve classification.

For the answer boundary detection task, we find that the multi-task setup (i.e., the classification layer participates in the training process) does not help its performance.
Since the classifier and pointer layer shared the encoding process, we originally expected that classification information can help detect answer boundaries.
But this is not the case. We think this is also reasonable since distinguishing whether the question is answerable is mainly focusing on the interactions between the passage-question pair, so once the question is predicted as answerable or not, it has nothing to do with the answer boundaries. This is consistent with how human-beings do this classification task.

We also run the test over SQuAD 1.1 development test to evaluate the performance.
Due to a condensed structure, our model achieves an $F1^*$ score of less than 86\%, which is not a very competitive score on SQuAD 1.1 test.  But as shown above, our model achieves a good score in SQuAD 2.0 test, which shows this model has the potential to achieve higher performance by making progress on both the answer detection and classification tasks.

Overall, we can conclude that our multi-task model works well since the performance of unanswerability classification improves significantly when the answer pointer and answer verifier work simultaneously.

\subsection{Study on the Different Thresholds of Unanswerability Classification}
\label{sub: More }

The output $b$ of the answer verifier is the probability of a question being unanswerable. The smaller the output, the lower the probability of unanswerability is. In SQuAD 2.0, the proportions of unanswerable questions are different in the training and test sets. The default threshold $0.5$ is optimized on the training set, but not suitable for the test set. Therefore,
it is reasonable to set a proper threshold to manually adapt to the test set.

As mentioned in SQuAD 2.0 paper \cite{squad2.0}, different thresholds for answerability prediction result in fluctuated scores between answerable and unanswerable questions.
Here we show the variation of the F1 score with different thresholds in Figure \label{fig:F1-t}.
The threshold between $[0,1]$ is used to decide whether a question can be answered. When the threshold is set to $0$, all questions are considered as answerable.

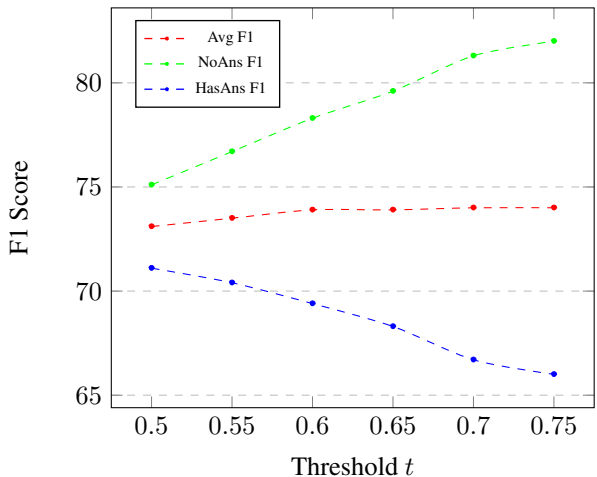
\begin{figure}[h]
    \centering
    \pgfplotsset{width=0.45\textwidth}
    \begin{tikzpicture}
      \begin{axis}[
      xlabel={Threshold $t$},
      ylabel={F1 Score},
      legend entries={Avg F1, NoAns F1, HasAns F1},
      mark size=1.0pt,
      ymajorgrids=true,
      grid style=dashed,
      legend pos= south east,
      legend style={font=\tiny,line width=.5pt,mark size=.5pt,
              at={(0.35,0.75)},
              /tikz/every even column/.append style={column sep=0.5em}},
              smooth,
      ]
      \addplot [red,dashed,mark=*] table [x index=0, y index=1] {L-F1.txt};
        \addplot [green,dashed,mark=*] table [x index=0, y index=3] {L-F1.txt};
      \addplot [blue,dashed,mark=*] table [x index=0, y index=2] {L-F1.txt};

      \end{axis}
  \end{tikzpicture}
  \caption{F1 score variation with different thresholds. ``NoAns F1'' is the recall of unanswerable questions.}\label{fig:F1-t}
  \end{figure}

As we can see, when the threshold is set to 0.5, F1 score of answerable questions is similar to that of unanswerable questions. When we increase the threshold (i.e., more likely to predict the question as unanswerable), performance for answerable questions degrades, and improves for unanswerable questions. This is as expected. We can see that the overall $F1$ score is slightly better, which is consistent with the idea from SQuAD 2.0. In addition, we find that for larger thresholds, the variance between $EM$ and $F1$ is narrowed since $EM$ and $F1$ scores for unanswerable questions are the same.

Finally, we set the threshold to be $0.7$ for the submission system to SQuAD evaluation.

\section{Related Work}
\subsection{End-to-end Models for MRC}
 Currently, end-to-end neural network models have achieved great successes for machine reading comprehension~\cite{seo2016bidirectional,kumar2015ask,sukhbaatar2015end,cui2016attention,xiong2016dynamic,dhingra2016gated,shen2016reasonet,hu2017reinforced,wang2018multi}.  Most of these models consist of three components: encoder, interaction, and pointer. The BiLSTM is widely used for encoding the embedded representation. For the interaction, bidirectional attention mechanism is very effective to fuse information of the question and passage. Finally, a pointer network \cite{pointernet} is used to predict the span boundaries of the answer. Specifically, in SQuAD test \cite{rajpurkar2016squad}, there are approaches to combine match-LSTM and pointer networks to produce boundaries of the answer and employ variant bidirectional attention mechanism to match the question and passage mutually.

 In our model, we learn from previous work and develop a condensed end-to-end model for the SQuAD 2.0 task.
 Different from the previous models, we use a unified representation to encode the question and passage simultaneously, and introduce a universal node to encode the fused information of the question and passage, which also plays an important role to predict the unanswerability of a question.

 \subsection{MRC with Unanswerable Questions}

 MRC with unanswerable questions is a more challenging task. Previous work \citet{levy2017zero,clark2017simple} has attempted to normalize a no-answer score depending on the probability of all answer spans and still detect boundaries at the same time.  But the scores of the answer span predictions are not very discriminative in distinguishing whether the question is answerable. Therefore, this kind of approaches, though relatively simple, cannot effectively deal with the answerability of a question.

 \citet{hu2018read,tan2018know} introduced an answer verifier idea to construct a classification layer.
 However, this kind of approaches usually has a pipeline structure. The answer pointer and answer verifier have their respective models that are trained separately.

 \paragraph{Multi-task models}
 Different from existing work, we regard the MRC with unanswerable questions as a multi-task learning problem \cite{multiDBLP:journals/ml/Caruana97} by sharing some meta-knowledge. Intuitively, answer prediction and answer verification are related tasks since the underlying comprehension and reasoning of language for these components is the same.
 Therefore, we construct a multi-task model to solve three sub-tasks: answer pointer, no-answer pointer, and answer verifier.

\section{Conclusion and Future Work }

In this paper, we regard the MRC with unanswerable questions as multi-task learning problems and propose the U-Net, a simple end-to-end model for MRC challenges. U-Net has good performance on SQuAD 2.0.  We first add a universal node to learn a fused representation from both the question and passage, then use a concatenated representation to pass through encoding layers. We only treat question and passage differently during attention interactions. In the rest blocks of U-Net, we still use the unified representation containing both the question and passage representation. Finally, we train the U-Net as a multi-task framework to determine the final answer boundaries as well as whether the question is answerable.  Our model has very simple structure yet achieves good results on SQuAD 2.0 test.

Our future work is to reconstruct the structure of U-Net by replacing the current multi-level attention block with a simpler self-attention mechanism, which we believe can capture the question and passage information, and intuitively is also coherent with the rest of our U-Net model.
In addition, we will improve the answer boundary detection performance based on some of the previous successful models.
Since our model actually does not achieve very competitive performance in the boundary detection task yet still has a good overall performance on SQuAD 2.0 test, we are optimistic that our U-Net model is potentially capable of achieving better performance.
Furthermore, our model has a simple structure and is easy to implement, therefore we believe that our model can be easily modified for various datasets.

\section{Acknowledgement }
 We would like to thank Robin Jia, Pranav Rajpurkar for their help with SQuAD 2.0 submissions.

\bibliographystyle{aaai}
\bibliography{nlp}

\begin{thebibliography}{}

\bibitem[\protect\citeauthoryear{{Bahdanau}, {Cho}, and
  {Bengio}}{2014}]{bahdanau2014neural}
{Bahdanau}, D.; {Cho}, K.; and {Bengio}, Y.
\newblock 2014.
\newblock Neural machine translation by jointly learning to align and
  translate.
\newblock {\em ArXiv e-prints}.

\bibitem[\protect\citeauthoryear{Caruana}{1997}]{multiDBLP:journals/ml/Caruana97}
Caruana, R.
\newblock 1997.
\newblock Multitask learning.
\newblock {\em Machine Learning} 28(1):41--75.

\bibitem[\protect\citeauthoryear{Chen \bgroup et al\mbox.\egroup }{2017}]{drqa}
Chen, D.; Fisch, A.; Weston, J.; and Bordes, A.
\newblock 2017.
\newblock Reading wikipedia to answer open-domain questions.
\newblock {\em CoRR} abs/1704.00051.

\bibitem[\protect\citeauthoryear{Clark and Gardner}{2017}]{clark2017simple}
Clark, C., and Gardner, M.
\newblock 2017.
\newblock Simple and effective multi-paragraph reading comprehension.
\newblock {\em arXiv preprint arXiv:1710.10723}.

\bibitem[\protect\citeauthoryear{Cui \bgroup et al\mbox.\egroup
  }{2016}]{cui2016attention}
Cui, Y.; Chen, Z.; Wei, S.; Wang, S.; Liu, T.; and Hu, G.
\newblock 2016.
\newblock Attention-over-attention neural networks for reading comprehension.
\newblock {\em arXiv preprint arXiv:1607.04423}.

\bibitem[\protect\citeauthoryear{Dhingra \bgroup et al\mbox.\egroup
  }{2016}]{dhingra2016gated}
Dhingra, B.; Liu, H.; Cohen, W.~W.; and Salakhutdinov, R.
\newblock 2016.
\newblock Gated-attention readers for text comprehension.
\newblock {\em arXiv preprint arXiv:1606.01549}.

\bibitem[\protect\citeauthoryear{Hermann \bgroup et al\mbox.\egroup
  }{2015}]{hermann2015teaching}
Hermann, K.~M.; Kocisky, T.; Grefenstette, E.; Espeholt, L.; Kay, W.; Suleyman,
  M.; and Blunsom, P.
\newblock 2015.
\newblock Teaching machines to read and comprehend.
\newblock In {\em Advances in Neural Information Processing Systems},
  1684--1692.

\bibitem[\protect\citeauthoryear{Hill \bgroup et al\mbox.\egroup
  }{2015}]{hill2015goldilocks}
Hill, F.; Bordes, A.; Chopra, S.; and Weston, J.
\newblock 2015.
\newblock The goldilocks principle: Reading children's books with explicit
  memory representations.
\newblock {\em arXiv preprint arXiv:1511.02301}.

\bibitem[\protect\citeauthoryear{Hochreiter and
  Schmidhuber}{1997}]{hochreiter1997long}
Hochreiter, S., and Schmidhuber, J.
\newblock 1997.
\newblock Long short-term memory.
\newblock {\em Neural computation} 9(8):1735--1780.

\bibitem[\protect\citeauthoryear{Hu \bgroup et al\mbox.\egroup
  }{2017}]{hu2017reinforced}
Hu, M.; Peng, Y.; Huang, Z.; Qiu, X.; Wei, F.; and Zhou, M.
\newblock 2017.
\newblock Reinforced mnemonic reader for machine reading comprehension.
\newblock {\em arXiv preprint arXiv:1705.02798}.

\bibitem[\protect\citeauthoryear{Hu \bgroup et al\mbox.\egroup
  }{2018}]{hu2018read}
Hu, M.; Peng, Y.; Huang, Z.; Yang, N.; Zhou, M.; et~al.
\newblock 2018.
\newblock Read+ verify: Machine reading comprehension with unanswerable
  questions.
\newblock {\em arXiv preprint arXiv:1808.05759}.

\bibitem[\protect\citeauthoryear{Huang \bgroup et al\mbox.\egroup
  }{2017}]{fusionnet}
Huang, H.; Zhu, C.; Shen, Y.; and Chen, W.
\newblock 2017.
\newblock Fusionnet: Fusing via fully-aware attention with application to
  machine comprehension.
\newblock {\em CoRR} abs/1711.07341.

\bibitem[\protect\citeauthoryear{Huang, Liu, and
  Weinberger}{2016}]{densenetDBLP:journals/corr/HuangLW16a}
Huang, G.; Liu, Z.; and Weinberger, K.~Q.
\newblock 2016.
\newblock Densely connected convolutional networks.
\newblock {\em CoRR} abs/1608.06993.

\bibitem[\protect\citeauthoryear{Kingma and
  Ba}{2014}]{admaDBLP:journals/corr/KingmaB14}
Kingma, D.~P., and Ba, J.
\newblock 2014.
\newblock Adam: {A} method for stochastic optimization.
\newblock {\em CoRR} abs/1412.6980.

\bibitem[\protect\citeauthoryear{Kumar \bgroup et al\mbox.\egroup
  }{2015}]{kumar2015ask}
Kumar, A.; Irsoy, O.; Su, J.; Bradbury, J.; English, R.; Pierce, B.; Ondruska,
  P.; Gulrajani, I.; and Socher, R.
\newblock 2015.
\newblock Ask me anything: Dynamic memory networks for natural language
  processing.
\newblock {\em arXiv preprint arXiv:1506.07285}.

\bibitem[\protect\citeauthoryear{Lee \bgroup et al\mbox.\egroup
  }{2016}]{featDBLP:journals/corr/LeeKP016}
Lee, K.; Kwiatkowski, T.; Parikh, A.~P.; and Das, D.
\newblock 2016.
\newblock Learning recurrent span representations for extractive question
  answering.
\newblock {\em CoRR} abs/1611.01436.

\bibitem[\protect\citeauthoryear{Levy \bgroup et al\mbox.\egroup
  }{2017}]{levy2017zero}
Levy, O.; Seo, M.; Choi, E.; and Zettlemoyer, L.
\newblock 2017.
\newblock Zero-shot relation extraction via reading comprehension.
\newblock {\em arXiv preprint arXiv:1706.04115}.

\bibitem[\protect\citeauthoryear{Liu \bgroup et al\mbox.\egroup
  }{2017}]{sanDBLP:journals/corr/abs-1712-03556}
Liu, X.; Shen, Y.; Duh, K.; and Gao, J.
\newblock 2017.
\newblock Stochastic answer networks for machine reading comprehension.
\newblock {\em CoRR} abs/1712.03556.

\bibitem[\protect\citeauthoryear{Pennington, Socher, and
  Manning}{2014}]{pennington2014glove}
Pennington, J.; Socher, R.; and Manning, C.~D.
\newblock 2014.
\newblock Glove: Global vectors for word representation.
\newblock {\em Proceedings of the Empiricial Methods in Natural Language
  Processing (EMNLP 2014)} 12:1532--1543.

\bibitem[\protect\citeauthoryear{Peters \bgroup et al\mbox.\egroup
  }{2018}]{elmo_}
Peters, M.~E.; Neumann, M.; Iyyer, M.; Gardner, M.; Clark, C.; Lee, K.; and
  Zettlemoyer, L.
\newblock 2018.
\newblock Deep contextualized word representations.
\newblock In {\em Proc. of NAACL}.

\bibitem[\protect\citeauthoryear{Rajpurkar \bgroup et al\mbox.\egroup
  }{2016}]{rajpurkar2016squad}
Rajpurkar, P.; Zhang, J.; Lopyrev, K.; and Liang, P.
\newblock 2016.
\newblock {SQuAD}: 100,000+ questions for machine comprehension of text.
\newblock {\em arXiv preprint arXiv:1606.05250}.

\bibitem[\protect\citeauthoryear{{Rajpurkar}, {Jia}, and
  {Liang}}{2018}]{squad2.0}
{Rajpurkar}, P.; {Jia}, R.; and {Liang}, P.
\newblock 2018.
\newblock {Know What You Don't Know: Unanswerable Questions for SQuAD}.
\newblock {\em ArXiv e-prints}.

\bibitem[\protect\citeauthoryear{Seo \bgroup et al\mbox.\egroup
  }{2016}]{seo2016bidirectional}
Seo, M.; Kembhavi, A.; Farhadi, A.; and Hajishirzi, H.
\newblock 2016.
\newblock Bidirectional attention flow for machine comprehension.
\newblock {\em arXiv preprint arXiv:1611.01603}.

\bibitem[\protect\citeauthoryear{Shen \bgroup et al\mbox.\egroup
  }{2016}]{shen2016reasonet}
Shen, Y.; Huang, P.-S.; Gao, J.; and Chen, W.
\newblock 2016.
\newblock Reasonet: Learning to stop reading in machine comprehension.
\newblock {\em arXiv preprint arXiv:1609.05284}.

\bibitem[\protect\citeauthoryear{Srivastava \bgroup et al\mbox.\egroup
  }{2014}]{dropoutDBLP:journals/jmlr/SrivastavaHKSS14}
Srivastava, N.; Hinton, G.~E.; Krizhevsky, A.; Sutskever, I.; and
  Salakhutdinov, R.
\newblock 2014.
\newblock Dropout: a simple way to prevent neural networks from overfitting.
\newblock {\em Journal of Machine Learning Research} 15(1):1929--1958.

\bibitem[\protect\citeauthoryear{Sukhbaatar \bgroup et al\mbox.\egroup
  }{2015}]{sukhbaatar2015end}
Sukhbaatar, S.; Weston, J.; Fergus, R.; et~al.
\newblock 2015.
\newblock End-to-end memory networks.
\newblock In {\em Advances in Neural Information Processing Systems},
  2431--2439.

\bibitem[\protect\citeauthoryear{Tan \bgroup et al\mbox.\egroup
  }{2018}]{tan2018know}
Tan, C.; Wei, F.; Zhou, Q.; Yang, N.; Lv, W.; and Zhou, M.
\newblock 2018.
\newblock I know there is no answer: Modeling answer validation for machine
  reading comprehension.
\newblock In {\em CCF International Conference on Natural Language Processing
  and Chinese Computing},  85--97.
\newblock Springer.

\bibitem[\protect\citeauthoryear{Vaswani \bgroup et al\mbox.\egroup
  }{2017}]{attentionisallyouneed}
Vaswani, A.; Shazeer, N.; Parmar, N.; Uszkoreit, J.; Jones, L.; Gomez, A.~N.;
  Kaiser, L.; and Polosukhin, I.
\newblock 2017.
\newblock Attention is all you need.
\newblock {\em CoRR} abs/1706.03762.

\bibitem[\protect\citeauthoryear{{Vinyals}, {Fortunato}, and
  {Jaitly}}{2015}]{pointernet}
{Vinyals}, O.; {Fortunato}, M.; and {Jaitly}, N.
\newblock 2015.
\newblock {Pointer Networks}.
\newblock {\em ArXiv e-prints}.

\bibitem[\protect\citeauthoryear{Wang and
  Jiang}{2016}]{matchDBLP:journals/corr/WangJ16a}
Wang, S., and Jiang, J.
\newblock 2016.
\newblock Machine comprehension using match-lstm and answer pointer.
\newblock {\em CoRR} abs/1608.07905.

\bibitem[\protect\citeauthoryear{Wang \bgroup et al\mbox.\egroup
  }{2017}]{selfmatchDBLP:conf/acl/WangYWCZ17}
Wang, W.; Yang, N.; Wei, F.; Chang, B.; and Zhou, M.
\newblock 2017.
\newblock Gated self-matching networks for reading comprehension and question
  answering.
\newblock In {\em Proceedings of the 55th Annual Meeting of the Association for
  Computational Linguistics, {ACL} 2017, Vancouver, Canada, July 30 - August 4,
  Volume 1: Long Papers},  189--198.

\bibitem[\protect\citeauthoryear{Wang, Yan, and Wu}{2018}]{wang2018multi}
Wang, W.; Yan, M.; and Wu, C.
\newblock 2018.
\newblock Multi-granularity hierarchical attention fusion networks for reading
  comprehension and question answering.
\newblock In {\em Proceedings of the 56th Annual Meeting of the Association for
  Computational Linguistics (Volume 1: Long Papers)}, volume~1,  1705--1714.

\bibitem[\protect\citeauthoryear{Xiong, Zhong, and
  Socher}{2016}]{xiong2016dynamic}
Xiong, C.; Zhong, V.; and Socher, R.
\newblock 2016.
\newblock Dynamic coattention networks for question answering.
\newblock {\em arXiv preprint arXiv:1611.01604}.

\end{thebibliography}

\end{document}